\documentclass{article}

\usepackage[preprint]{neurips_2025}

\usepackage[utf8]{inputenc} 
\usepackage[T1]{fontenc}    
\usepackage{hyperref}       
\usepackage{url}            
\usepackage{booktabs}       
\usepackage{amsfonts}       
\usepackage{nicefrac}       
\usepackage{microtype}      
\usepackage{xcolor}         

\usepackage{colortbl}
\usepackage{color}
\usepackage{bbding}
\usepackage{multirow}
\usepackage{courier}
\usepackage{pifont}
\usepackage{bbm}
\usepackage{dsfont}

\usepackage{graphicx}
\usepackage{amsmath}
\usepackage{wrapfig} 

\title{TextFlux: An OCR-Free DiT Model for High-Fidelity Multilingual Scene Text Synthesis}

\author{%
  Yu Xie$^{1}$\thanks{Equal contribution.}, Jielei Zhang$^{1  *}$, Pengyu Chen$^{1}$,  Weihang Wang$^{1}$, Longwen Gao$^{1}$  \\
  {\bfseries Peiyi Li$^{1}$, Qian Qiao$^{1,2}$, Zhouhui Lian$^{3}$}\thanks{Corresponding author} \\
  $^{1}$bilibili Inc. \! $^{2}$OpenWPLab \! $^{3}$Wangxuan Institute of Computer Technology, Peking University \\
  \texttt{\{xieyu20001003,yctmzjl\}@gmail.com}, 
  \texttt{lianzhouhui@pku.edu.cn} \\
  \url{https://yyyyyxie.github.io/textflux-site/}
}

\begin{document}

\maketitle

\begin{figure}[ht]
\vspace{-26pt}
\begin{center}
	\includegraphics[width=0.97\linewidth]{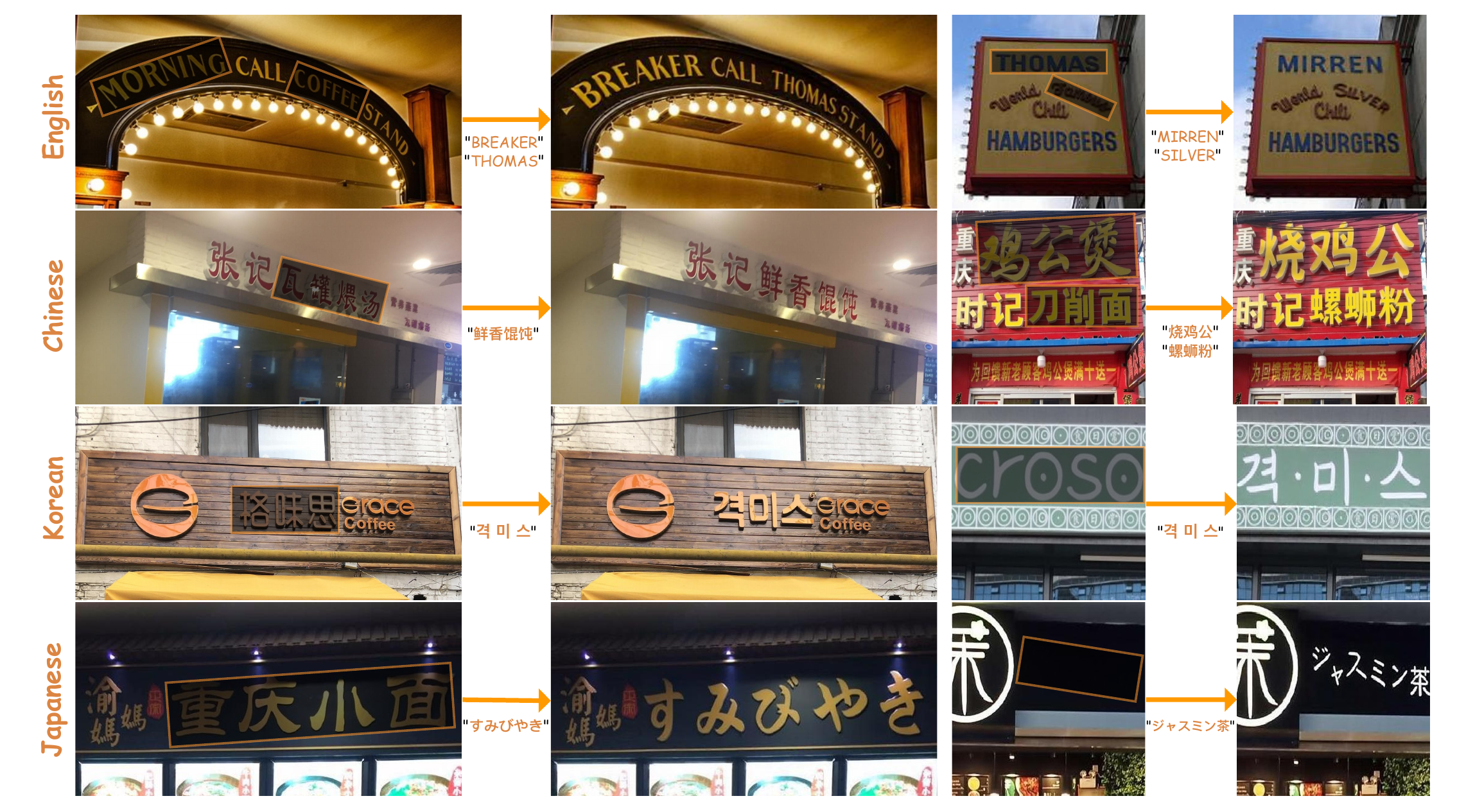}
\end{center}
\vspace{-9pt}
\caption{\small Some examples of high-fidelity multilingual scene text images generated by our TextFlux.
}
\vspace{-4pt}
\label{fig:abs}
\end{figure}

\vspace{-2pt} 
\begin{abstract}
  Diffusion-based scene text synthesis has progressed rapidly, yet existing methods commonly rely on additional visual conditioning modules and require large-scale annotated data to support multilingual generation. 
In this work, we revisit the necessity of complex auxiliary modules and further explore an approach that simultaneously ensures glyph accuracy and achieves high-fidelity scene integration, by leveraging diffusion models' inherent capabilities for contextual reasoning.
To this end, we introduce TextFlux, a DiT-based framework that enables multilingual scene text synthesis.
The advantages of TextFlux can be summarized as follows:
(1) OCR-free model architecture. TextFlux eliminates the need for OCR encoders (additional visual conditioning modules) that are specifically used to extract visual text-related features.
(2) Strong multilingual scalability. TextFlux is effective in low-resource multilingual settings, and achieves strong performance in newly added languages with fewer than 1,000 samples.
(3) Streamlined training setup. TextFlux is trained with only 1\% of the training data required by competing methods.
(4) Controllable multi-line text generation. 
TextFlux offers flexible multi-line synthesis with precise line-level control, outperforming methods restricted to single-line or rigid layouts.
Extensive experiments and visualizations demonstrate that TextFlux outperforms previous methods in both qualitative and quantitative evaluations.
\end{abstract}

\begin{figure}[thbp] 
\centering                
\includegraphics[width=1.0\textwidth]{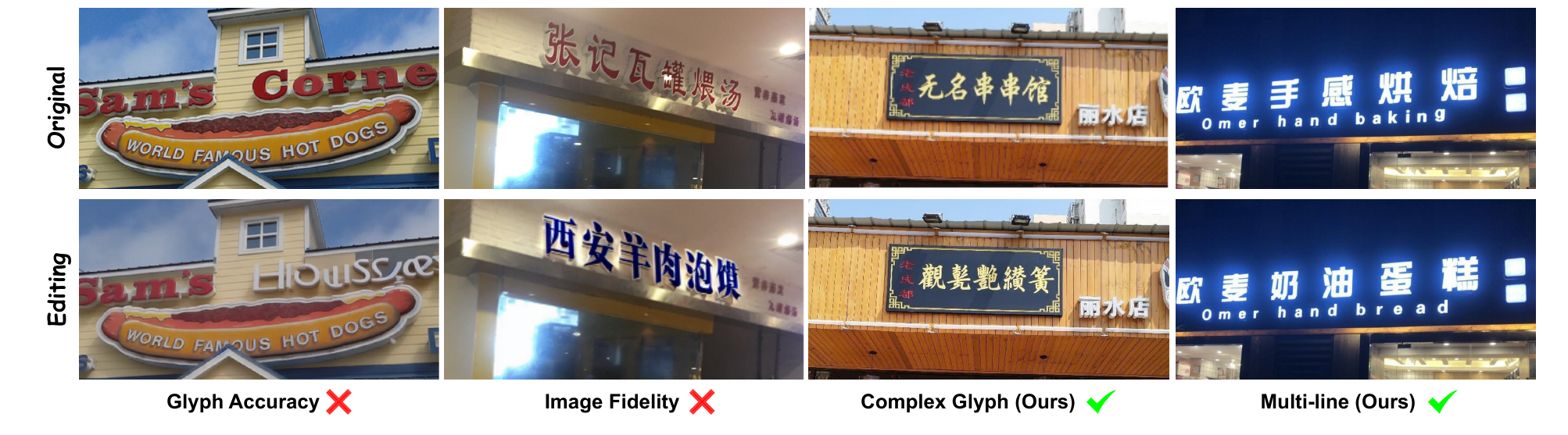}
\caption{TextFlux addresses the common conflict between glyph accuracy and stylistic integration in scene text synthesis. Prior works often exhibit either glyph errors (first column) or poor visual fidelity and integration (second column). In contrast, TextFlux accurately renders complex and multi-line text with high fidelity to the scene context (third and fourth columns).}\label{error_type}  
\end{figure}

\section{Introduction}
\label{sec:introduction}

The synthesis of scene text in this work encompasses both \textit{text reconstruction} and \textit{text editing}, aiming to restore or modify textual content in natural images while preserving the visual fidelity of the scene.
The challenges of this task can be categorized into two core aspects: first, ensuring the \textbf{``spelling'' accuracy} of the generated text itself -- that is, the correctness of its glyph structure; and second, \textbf{naturally and realistically} integrating the edited or generated text into the complex visual contexts of diverse target scenes.

To address the first core challenge (ensuring the accuracy of the glyph structure), existing methods~\cite{chen2024textdiffuser,tuo2023anytext,zhao2023udifftext,yang2023glyphcontrol,ma2024glyphdraw2}  often introduce \textit{specialized textual features} (such as explicit glyph information) as conditions. However, while leveraging such specialized textual features for strong, specific control does improve the accuracy of the generated glyphs, 
it tends to cause the generated text to appear merely ``pasted on'' and lack realistic integration with the scene, as shown in Fig.\ref{error_type}. To address this issue of overall visual fidelity (the second core challenge),
some approaches~\cite{tuo2024anytext2,wang2024high,zeng2024textctrl,du2025textcrafter} attempt to establish independent controls for distinct visual attributes such as style, font, and color, injecting corresponding features as conditions.
However, the inherent \textit{diversity, complexity, and subjectivity} of text visual styles make it extremely difficult to construct a comprehensive universal representation for them. Moreover, some attributes, such as lighting and texture, are inherently hard to disentangle, greatly increasing the complexity of model design and training.

Considering the aforementioned challenges, this paper aims to explore a new approach to reconcile the conflict between glyph accuracy and realistic integration in scene text synthesis. 
We observe that current diffusion models~\cite{rombach2021highresolution,podell2023sdxl,peebles2023scalable,blackforestlabs_flux} already excel in maintaining overall contextual coherence and visual fidelity in inpainting tasks. The real challenge lies in enabling them to \textbf{``learn to spell'' from scratch}, especially for complex character systems like Chinese with its intricate strokes. If the model inherently knew the specific details of glyph structures, it could theoretically generate text with high visual fidelity. Based on these insights, we depart from the traditional approach of feature-level conditioning and instead turn to the image's own spatial dimension: by directly providing a visual glyph reference, we transform the core task from ``learning to spell'' to \textbf{learning how to integrate this given glyph into the context with a scene-adaptive style}.
This simplified learning objective allows the model to focus on the integration process by leveraging its inherent strengths, rather than on the complex task of ``learning to spell'' from scratch.

In this paper, we propose TextFlux, an OCR-free diffusion framework for multi-language scene text synthesis, built upon the state-of-the-art DiT-based Flux architecture~\cite{blackforestlabs_flux}. TextFlux guides the model to adaptively infer and render harmonious text styles from the scene context. This approach circumvents the dilemmas faced by existing methods in the definition and control of various text visual attributes, offering a concise and efficient solution for generating high-fidelity, contextually consistent text. Furthermore, benefiting from the design of this new paradigm, TextFlux demonstrates strong capabilities in simultaneously editing multi-line text, handling multiple languages, rendering complex glyphs, and even achieving zero-shot generalization to characters not seen in the training set.

Our main contributions can be summarized as follows:
\begin{itemize}
\item We propose TextFlux, an OCR-free diffusion framework for scene text synthesis. TextFlux introduces essential textual guidance by spatially integrating glyph-rendered visual cues, thereby eliminating the need for dedicated OCR encoders for various visual text attributes.
\item We demonstrate that TextFlux achieves strong multilingual scalability, especially in low-resource languages, effectively synthesizing text across multiple languages and rapidly adapting to new, low-resource languages with minimal language-specific data.
\item We enable flexible and controllable multi-line text generation through inherent spatial guidance, allowing precise line-level control over content and position. Extensive experiments on multiple benchmarks demonstrate that TextFlux achieves state-of-the-art performance in multilingual scene text synthesis, outperforming existing methods in both visual fidelity and sequence accuracy.
\end{itemize}

\section{Related Work}

\subsection{Text-to-Image Synthesis}
In recent years, diffusion models have achieved significant success across various tasks, especially in text-to-image synthesis~\cite{dhariwal2021diffusion,rombach2021highresolution,mou2024t2i,liu2024glyph2,hu2024amo,gong2025seedream}, image-to-image translation~\cite{saharia2022palette}, and image editing~\cite{chen2025edit,brooks2023instructpix2pix,hertz2022prompt,feng2024dit4edit}. These successes demonstrate the superiority of diffusion models in the field of image generation. 
Emerged areas of exploration include Personalized Generation~\cite{ruiz2023dreambooth,ruiz2024hyperdreambooth,chen2025posta}, Controllable text-to-image (T2I) Generation~\cite{zhang2023adding,li2024controlnet++}, LLM-assisted T2I~\cite{feng2023layoutgpt}, Style Transfer~\cite{wu2021styleformer}, and Safety Issues~\cite{li2024self,wu2024universal}. To further enhance generation performance, recent studies integrate large-scale transformer architectures as the backbone of diffusion models, resulting in advanced models like DiT~\cite{peebles2023scalable,blackforestlabs_flux,chen2023pixart}. 
Among these architectural innovations, Flux~\cite{blackforestlabs_flux}, which is based on flow matching objectives~\cite{lipman2022flow}, has achieved state-of-the-art generation results and has been open-sourced. These advancements have subsequently fueled research into the control~\cite{tan2024ominicontrol,yu2024representation} and acceleration~\cite{tan2025ominicontrol2,ma2025inference} of these new architectures.

\subsection{Scene Text Synthesis}
Despite the rapid development of diffusion models, these general methods often face limitations when generating scene text. Early researchers pointed out that text encoders play a crucial role in generating accurate text. To address this issue, Imagen~\cite{saharia2022photorealistic}, eDiff-I~\cite{balaji2022ediff}, and DeepFloyd~\cite{deepfloyd} utilized large-scale language models (e.g., T5-XXL~\cite{chung2024scaling}) to optimize text spelling capabilities.  UDiffText~\cite{zhao2023udifftext} attempts to train a text encoder aligned with visual text features to replace the text encoder in CLIP~\cite{radford2021learning}, thereby enhancing the glyph-awareness. However, improvements on text encoders bring only limited gains in text rendering quality within diffusion models, especially for non-Latin scripts.





As a result, more scene text synthesis methods~\cite{zhang2024brush,tuo2023anytext,ma2024glyphdraw2,wang2025glyphmastero,gao2025postermaker} are focused on designing specialized condition control modules specifically tailored to visual text. GlyphDraw~\cite{ma2023glyphdraw} initially used glyph images as condition control and rendered characters at the center. GlyphControl~\cite{yang2023glyphcontrol} further extended this approach by spatially aligning the glyph rendering position with the actual text generation position. TextDiffuser~\cite{chen2023textdiffuser} trained an additional OCR engine to generate segmentation masks, which are used for condition control. AnyText~\cite{tuo2023anytext} inherited the design philosophy of condition control from GlyphControl and expanded it to multilingual versions. DreamText~\cite{wang2024high} further introduced additional control conditions such as different fonts to enhance the rendering capability of visual text. Besides these methods, some approaches~\cite{fang2025recognition,zeng2024textctrl} aim to reduce the difficulty of visual text editing by cropping the text to be edited and only processing text lines. Although these methods significantly improve character accuracy, they often sacrifice visual fidelity in text generation due to the lack of context integration with the entire image.

Although the various OCR encoders proposed by the aforementioned methods enhanced the effectiveness of scene text synthesis, they also led to architectural redundancy and optimization difficulties due to excessive condition control. Moreover, the overemphasis on OCR characteristics often results in a loss of fidelity. In the new wave of control and acceleration based on the latest DiT series of architectures~\cite{blackforestlabs_flux,peebles2023scalable}, this paper 
seeks to shift the paradigm away from using specialized condition control modules (OCR encoders) in scene text synthesis. Instead, it introduces a novel approach that leverages contextual information from the image itself to achieve scene-adaptive and visually coherent text generation.

\section{Methodology}
\subsection{ { Preliminary } }

While U-Net has been the dominant architecture in early diffusion models, recent works like FLUX-1~\cite{blackforestlabs_flux}, Stable Diffusion 3~\cite{rombach2021highresolution}, and PixArt~\cite{chen2023pixart} have explored the Transformer-based DiT architecture~\cite{peebles2023scalable}. These DiT models scale well to larger sizes and demonstrate an improved ability to understand the overall context and relationships within the images.
Notably, OmniControl~\cite{tan2024ominicontrol} and In-Context LoRA~\cite{huang2024context} further suggest that DiT-based architectures inherently possess contextual reasoning capabilities. 
These insights motivate a new perspective on control mechanisms specifically for scene text synthesis, where contextual understanding plays a key role.
Among the DiT-based architectures, FLUX-1-Fill-dev~\cite{blackforestlabs_flux} is an inpainting-oriented variant that supports flexible conditioning.
In this design, the standard DiT input—noisy image tokens \( \mathbf{X} \in \mathbb{R}^{N \times d} \) and text conditioning tokens \( \mathbf{T}_{c} \in \mathbb{R}^{M \times d} \)—is extended for the inpainting task by introducing masked image tokens \( \mathbf{X}_i \in \mathbb{R}^{N \times d} \) and binary mask tokens \( \mathbf{X}_m \in \mathbb{R}^{N \times d} \) resulting in an augmented visual sequence:
\begin{equation}
\mathbf{Z} = \text{Concat}(\{\mathbf{X}, \mathbf{X}_i, \mathbf{X}_m\}, \text{dim}=-1).
\end{equation}
The sequence \( \mathbf{Z} \), along with the text conditioning tokens \( \mathbf{T}_{c} \), is then fed into the DiT blocks. This architecture serves as the foundation of TextFlux.

\subsection{ {Motivation} }
\begin{wrapfigure}[18]{r}{0.49\textwidth}
\centering
\vspace{-8pt}
\includegraphics[width=0.49\textwidth]{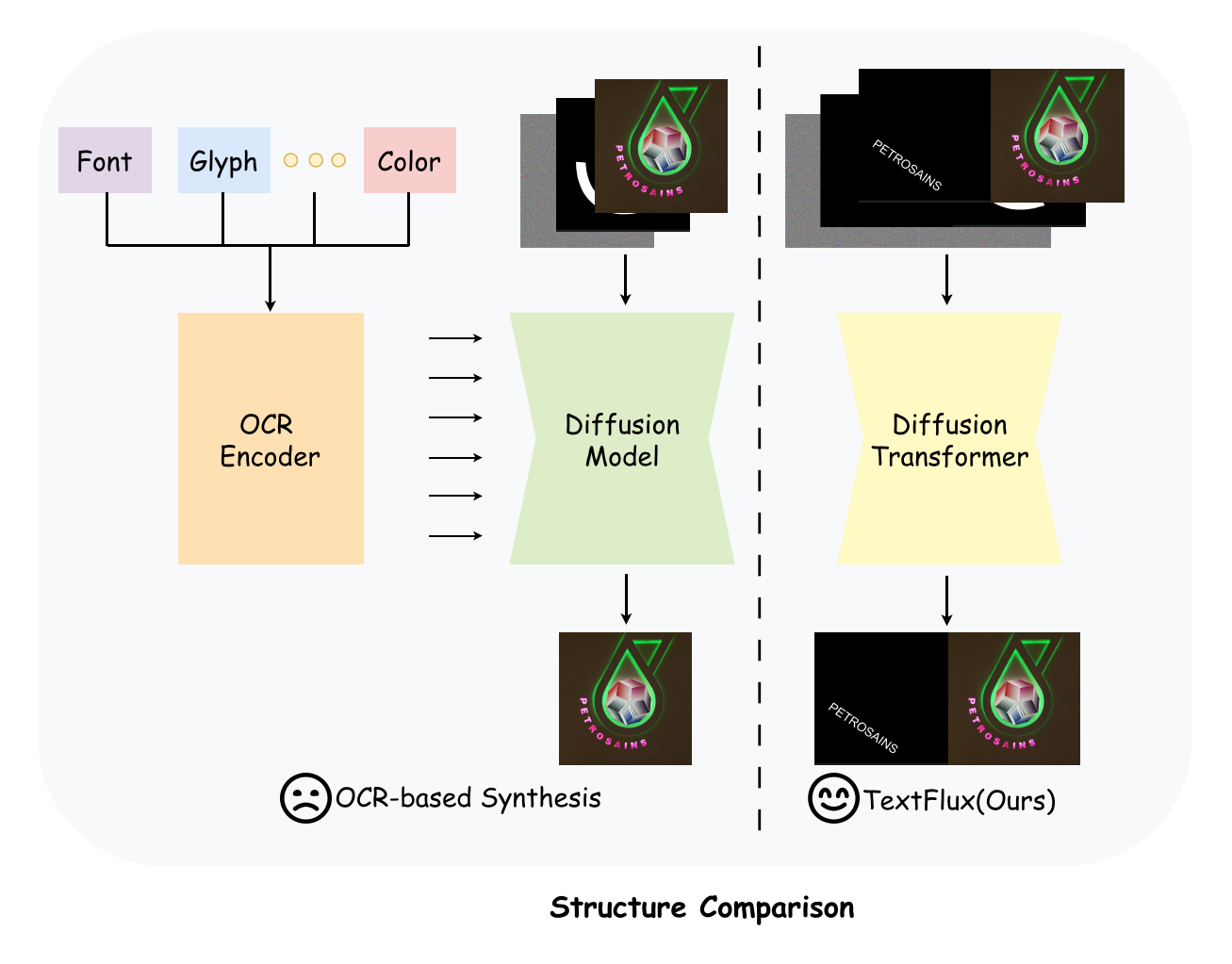}
\vspace{-12pt}
\caption{\small
Traditional methods employ OCR encoders to extract and inject various visual text features (e.g., font, glyph, color) as conditions. 
TextFlux streamlines the process by directly providing spatial glyph cues.
}
\label{fig:compare_vs_ours}
\end{wrapfigure}

Recent diffusion-based scene text synthesis methods typically employ additional visual conditioning modules named OCR encoders, as shown in Fig.~\ref{fig:compare_vs_ours} (left). Although the design of these approaches seems reasonable, they possess several critical limitations. First, integrating diverse OCR encoders considerably increases model architecture complexity. The text-specific feature representations may be alien to the general pre-trained diffusion model, necessitating extensive learning from scratch and complicating optimization. Second, the aforementioned optimization difficulty often demands large-scale annotated datasets~\cite{tuo2023anytext,chen2023textdiffuser} and prolonged training, hindering scalability, especially for low-resource languages. Third, the use of OCR-based modules typically leads to the requirement of additional specialized loss functions~\cite{chen2023textdiffuser,tuo2023anytext}, significantly increasing the implementation complexity and computational cost. 
Last but not least, the heavy reliance on these modules biases the model towards fitting the specific OCR representations. This could potentially lead the model to overlook the broader scene context, ultimately undermining the visual fidelity and natural integration of the synthesized text. It is similar to the ``pasted-on'' appearance discussed in the Introduction section.

In summary, the proposed TextFlux consists of the following advantages: 1) By eliminating the reliance on OCR encoders, both efficiency and architectural simplicity can be achieved. 2) Our training strategy focuses on enabling the diffusion model to adapt a provided glyph to the scene image context, which can be markedly simplified by our new paradigm. 3) We significantly lessen the dependency on large-scale annotated data, especially for multilingual settings. Thus, even in low-resource scenarios, excellent performance can be achieved with only minimal additional data (e.g., adapting to new languages).
A key insight serves as the foundation for our proposed TextFlux's simplified paradigm: pretrained diffusion transformers inherently possess strong capabilities for contextual reasoning and visual understanding, which we leverage by concatenating glyphs spatially.

\subsection{ { TextFlux } }

\begin{figure*}[t]
  \centering
  \includegraphics[width=1\textwidth]{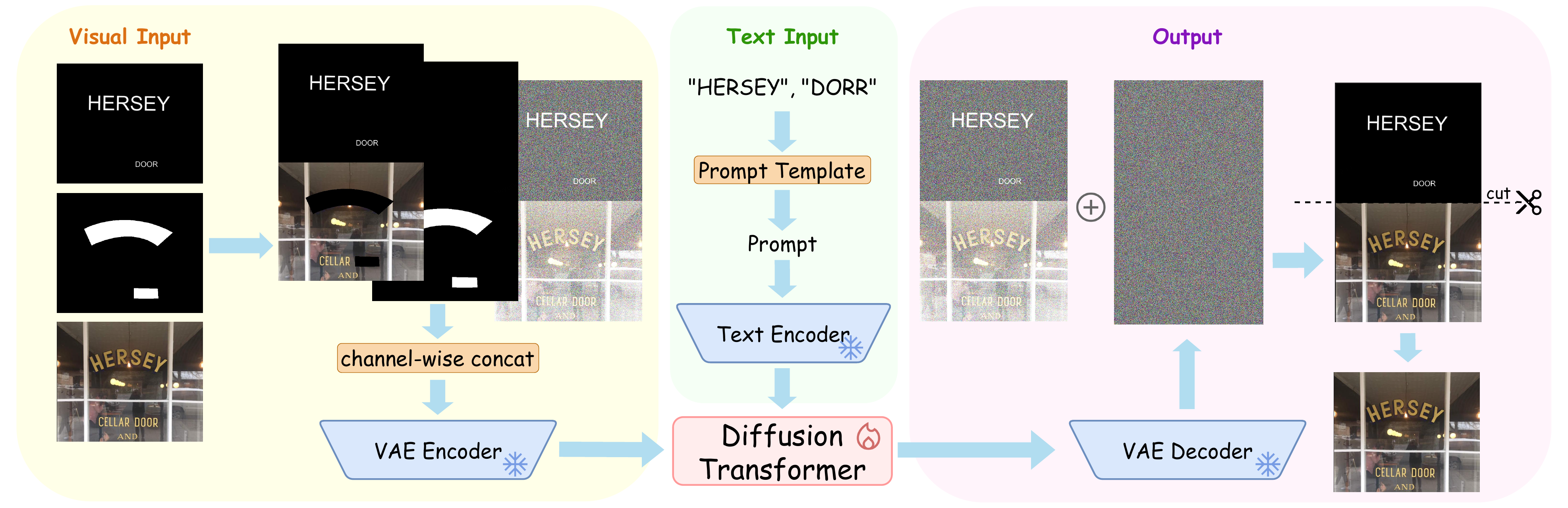}
  \caption{\textbf{Overview of TextFlux.} We propose an OCR-free scene text synthesis method that spatially concatenates glyph-rendered text with the original image as model input, enabling the diffusion transformer to leverage its inherent context-awareness to render text in the masked regions.}
  \label{arch}
\end{figure*}

Based on the above-mentioned analyses, we utilize FLUX.1-Fill-dev~\cite{blackforestlabs_flux}, an inpainting-oriented variant from the DiT family, to develop TextFlux, a scene text synthesis system that supports multilingual scenarios through an efficient concatenation scheme. The overall architecture is illustrated in Fig.~\ref{arch}. We describe the system from the perspective of input construction.


\noindent\textbf{Model Input.}
Our method prepares the glyph-guided image input for the diffusion model. First, the target text is rendered as white foreground on a black background to create a binary glyph mask \(\mathbf{I}_{\text{glyph}}\), ensuring it matches the resolution of the scene image \(\mathbf{I}_{\text{scene}}\). Second, \(\mathbf{I}_{\text{glyph}}\) is spatially concatenated with \(\mathbf{I}_{\text{scene}}\) (either horizontally or vertically, as determined by the chosen \(\textit{axis}\)) to form the combined input \(\mathbf{I}_{\text{concat}} = \operatorname{Concat}([\mathbf{I}_{\text{glyph}}, \mathbf{I}_{\text{scene}}], \textit{axis})\). This input structure enables the model to directly observe the precise glyph template alongside the full scene context.

\noindent\textbf{Prompt design.}  
Following the paradigm of in-context learning in diffusion models~\cite{huang2024context}, we additionally provide a descriptive text prompt
to accompany each input image. The prompt is designed to clarify the roles of the two concatenated images and the target text content. It follows the template:
\textit{"The pair of images highlights some white words on a black background, as well as their style on a real-world scene image. [IMAGE1] is a template image rendering the text, with the words \{words\}; [IMAGE2] shows the text content \{words\} naturally and correspondingly integrated into the image."}
Here, "\{words\}" is replaced by the actual text to be rendered. During training, this prompt guides the model to understand the semantic relationship between the glyph template and the scene image.



Consequently, by spatially concatenating \(\mathbf{I}_{\text{glyph}}\) and \(\mathbf{I}_{\text{scene}}\) into a unified input \(\mathbf{I}_{\text{concat}}\), TextFlux offers a direct and information-rich visual guidance mechanism. This design enables the model to concentrate on its well-developed pre-trained capabilities for contextual understanding and visual fusion, facilitating the efficient synthesis of high-quality scene text.





\subsection{{Model Training and Inference}}


To train the model, we adopt a flow-matching objective as introduced in the Flux framework~\cite{blackforestlabs_flux}. Given a clean latent representation \( \mathbf{x}_0 \), a noise vector \( \mathbf{z}_1 \sim \mathcal{N}(0, \mathbf{I}) \), and a noise scale \( \sigma_t \) associated with the random time step \( t \), the noisy latent input is generated by convex interpolation:
\begin{equation}
\mathbf{x}_t = (1 - \sigma_t)\, \mathbf{x}_0 + \sigma_t\, \mathbf{z}_1.
\end{equation}
The model is trained to predict the velocity between \( \mathbf{x}_0 \) and \( \mathbf{z}_1 \), with the training loss defined as:
\begin{equation}
\mathcal{L}_{\text{FM}} = \mathbb{E}_{t,\, \mathbf{x}_0,\, \mathbf{z}_1} \left[ \omega_t \cdot \left\| \hat{\mathbf{v}}_\theta(\mathbf{x}_t, t, \mathbf{c}) - (\mathbf{z}_1 - \mathbf{x}_0) \right\|_2^2 \right],
\end{equation}
where \( \hat{\mathbf{v}}_\theta \) is the model prediction, \( \omega_t \) is a time-dependent weighting factor, and \( \mathbf{c} \) includes the conditioning features such as the concatenated image, text prompt embeddings, and inpainting mask features. No additional perceptual loss is used, keeping the training objective simple and stable.




During the inference stage, as illustrated in Fig.~\ref{arch}, the user provides three inputs: a scene image to be edited, a binary mask indicating the target text region, and the desired text content. The pipeline automatically generates a glyph-based template image, concatenates it with the input scene image, and feeds the result into the model. The output image is cropped to remove the template region, resulting in the final edited scene image.

\subsection{Implementation Details}

Our method is built on the pre-trained FLUX.1-Fill-dev, a latent rectified flow transformer model for image synthesis. For training, we set the batch size to 1 and use the gradient accumulation of 8. We employ the AdamW optimizer with a constant learning rate of 2e-5, running for 30,000 iterations in total.  
Since resolution is critical for scene text tasks, we develop a specialized data augmentation approach by resizing the image’s longer side to 512, 640, 768, 896, or 1024, thus obtaining input images of various resolutions. During training, we directly mix data from different languages.  
We train two versions of TextFlux: the first one trained for its full parameters on two A100 (80 GB) GPUs, and the other one trained via LoRA on a single A100 (80 GB) GPU with a LoRA rank of 128.

\section{Experiment}

\subsection{Datasets and Evaluation Metrics}
\label{sec4-1}

\noindent\textbf{Datasets.}
In previous studies, large-scale datasets are commonly employed for multilingual visual text generation tasks. For instance, the AnyWord-3M~\cite{tuo2023anytext} dataset contains approximately three million publicly sourced multilingual images, while the MARIO-10M~\cite{chen2023textdiffuser} dataset comprises around ten million images that are primarily in English, though a small portion may include other languages. In contrast to these large-scale datasets, we use a relatively small training set of 30{,}405 images: approximately 10{,}000 in English, 15{,}000 in Chinese, and 1{,}000 each for Japanese, Korean, French, German, and Italian.
Specifically, the English data primarily come from MLT2017~\cite{MLT}, TotalText~\cite{ch2020total}, and CTW1500~\cite{ch2020total} training sets commonly used in OCR-related tasks~\cite{xie2024dntextspotter,ye2023deepsolo}; the Chinese data are mainly derived from the ReCTS~\cite{ReCTS} and RCTW~\cite{RCTW} training sets; the remaining languages are obtained from the MLT2019~\cite{MLT} competition data.

For validation, we use the test set provided in~\cite{tuo2023anytext} from the AnyWord-3M dataset, which includes 1,000 English and 1,000 Chinese images. To further evaluate our method under more challenging conditions, we additionally include two harder test sets: TotalText~\cite{ch2020total} test set for English, featuring 300 images with curved and arbitrarily shaped text, and the ReCTS~\cite{ReCTS} test set for Chinese, consisting of 2,000 real-world images with diverse and complex layouts. These datasets provide a more rigorous benchmark for assessing the robustness and generalization capability of our method, particularly under complex and diverse text conditions.

\noindent\textbf{Evaluation.} 
We evaluate our method on two tasks: scene text reconstruction and scene text editing. In scene text reconstruction, the text image is reconstructed by rendering text in the masked region using the words directly from the ground truth text labels. In scene text editing, the original words in the labels are replaced with a random word. 
For evaluation, we use off-the-shelf scene text recognition (STR) models to calculate recognition accuracy, primarily measured by Sentence Accuracy (Sen. Acc), with additional analysis using Normalized Edit Distance (NED) provided in the appendix.

To further evaluate the difference between synthetic and real images, we use Frechet Inception Distance (FID) and Learned Perceptual Image Patch Similarity (LPIPS) to assess the visual fidelity of the generated images. In addition, we conduct a user study, where participants are asked to rate the generated results on a scale from 1 to 10, based on overall visual quality and realism. The averaged user scores serve as a subjective evaluation to complement the quantitative metrics.

\begin{table*}[htb]
    \caption{\textbf{Quantitative comparison of multi-line text generation metrics against baselines.} We use Sequence Accuracy (SeqAcc) as the main evaluation metric to measure recognition correctness. The best scores are highlighted in bold. The second-best results are underlined. FID and LPIPS are computed on the ReCTS dataset. The User Study (US) is conducted to capture human evaluations regarding the overall quality of generated images (score range: 0--10). Detailed FID and LPIPS results on other datasets, as well as additional Normalized Edit Distance (NED) results, are provided in the appendix.}
    \centering
    \resizebox{\linewidth}{!}{
    \begin{tabular}{@{}l|cccc|cccc|c|c|c@{}}
     \toprule[1.5pt]
     \multirow{2}{*}{\textbf{Method}} 
       & \multicolumn{4}{c|}{\textbf{SeqAcc-Recon (\%)}$\uparrow$} 
       & \multicolumn{4}{c|}{\textbf{SeqAcc-Editing (\%)}$\uparrow$} 
       & \multirow{2}{*}{\textbf{FID}$\downarrow$} 
       & \multirow{2}{*}{\textbf{LPIPS}$\downarrow$} 
       & \multirow{2}{*}{\textbf{US}$\uparrow$}\\ 
     \cmidrule(lr){2-5} \cmidrule(lr){6-9}
     & \textbf{AnyWord(EN)} & \textbf{AnyWord(CH)} & \textbf{TotalText} & \textbf{ReCTS} 
       & \textbf{AnyWord(EN)} & \textbf{AnyWord(CH)} & \textbf{TotalText} & \textbf{ReCTS} 
       &  &  &  \\
     \midrule[1pt]
     Flux~\cite{blackforestlabs_flux} 
       & 43.0 & 9.3 & 29.5 & 4.8  
       & 11.6 & 0.0 & 11.5 & 0.0 
       & 18.25 & 0.1431 & 4.5\\
     AnyText~\cite{tuo2023anytext} 
       & 14.8 & 24.1 & 6.5 & 20.6 
       & 13.7 & 19.2 & 4.6 & 18.5 
       & 22.57 & 0.4095 & 3.8\\
     AnyText2~\cite{tuo2024anytext2} 
       & 23.7 & 28.1 & 15.5 & 25.2 
       & 17.0 & 24.2 & 15.0 & 23.6 
       & 21.75 & 0.3054 & 4.3\\
     \midrule[1pt]
     TextFlux(LoRA) 
       & \underline{76.7} & \underline{50.8} & \underline{62.3} & \underline{56.6} 
       & \underline{61.1} & \underline{32.8} & \underline{35.4} & \underline{32.1} 
       & \underline{12.09} & \underline{0.1038} & \underline{7.4}\\
     TextFlux 
       & \textbf{77.3} & \textbf{61.4} & \textbf{62.9} & \textbf{64.1} 
       & \textbf{63.8} & \textbf{40.7} & \textbf{36.2} & \textbf{37.2} 
       & \textbf{11.02} & \textbf{0.0975} & \textbf{8.0}\\
     \bottomrule[1.5pt]
    \end{tabular}}
    \label{tab:quan_results1}
\end{table*}

\begin{table*}[htb]
    \caption{\textbf{Quantitative comparison of single-line text generation metrics against baselines.}}
    \centering
    \resizebox{\linewidth}{!}{
    \begin{tabular}{@{}l|cccc|cccc@{}} 
      \toprule[1.5pt]
      \multirow{2}{*}{\textbf{Method}} 
        & \multicolumn{4}{c|}{\textbf{SeqAcc-Recon (\%)}$\uparrow$} 
        & \multicolumn{4}{c}{\textbf{SeqAcc-Editing (\%)}$\uparrow$} \\ 
      \cmidrule(lr){2-5} \cmidrule(lr){6-9}
      & \textbf{AnyWord(EN)} & \textbf{AnyWord(CH)} & \textbf{TotalText} & \textbf{ReCTS} 
        & \textbf{AnyWord(EN)} & \textbf{AnyWord(CH)} & \textbf{TotalText} & \textbf{ReCTS} \\ 
      \midrule[1pt]
      Flux~\cite{blackforestlabs_flux} 
        & 53.6 & 6.6 & 63.2 & 10.0  
        & 42.1 & 0.0 & 41.6 & 0.0 \\
      AnyText~\cite{tuo2023anytext} 
        & 34.8 & 31.7 & 11.4 & 36.2 
        & 30.9 & 28.4 & 10.5 & 30.4 \\
      AnyText2~\cite{tuo2024anytext2} 
        & 45.1 & 35.9 & 20.5 & 41.5 
        & 42.0 & \underline{37.5} & 21.3 & 34.6 \\
      \midrule[1pt]
      TextFlux(LoRA) 
        & \underline{80.1} & \underline{52.7} & \textbf{66.1} & \underline{63.4} 
        & \underline{55.5} & 36.8 & \textbf{45.0} & \underline{37.2} \\
      TextFlux 
        & \textbf{80.3} & \textbf{62.3} & \underline{65.3} & \textbf{68.5} 
        & \textbf{56.2} & \textbf{48.2} & \underline{41.9} & \textbf{40.6} \\
      \bottomrule[1.5pt]
    \end{tabular}}
    \label{tab:quan_results2} 
\end{table*}

\begin{figure}[thbp]
  \centering
  \includegraphics[width=1\textwidth]{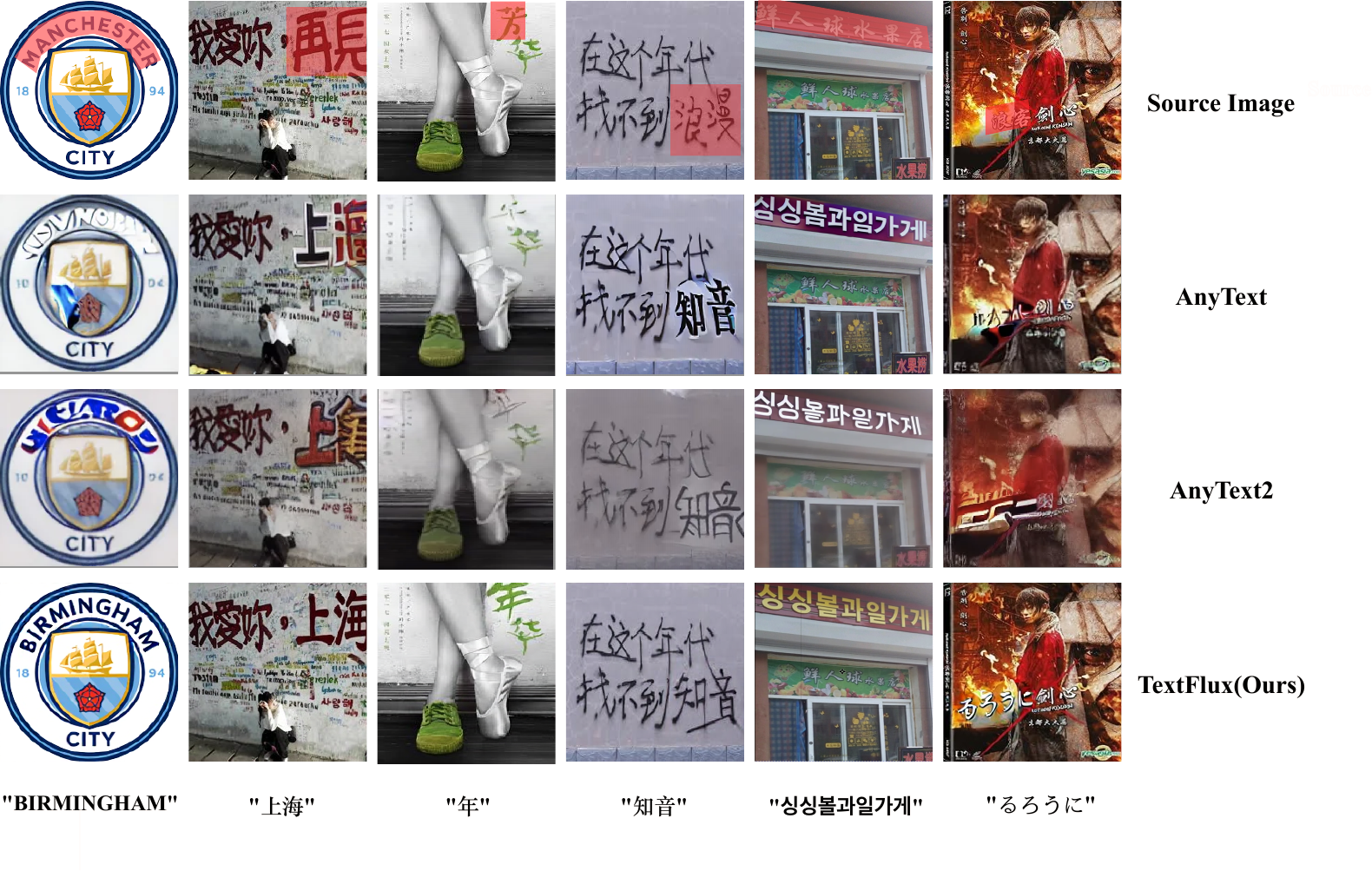}
  \caption{\textbf{Comparison of scene text synthesis methods:} AnyText, AnyText2, and our TextFlux. More results are available in the appendix.}
  \label{visualization}
\end{figure}

\subsection{Quantitative and Qualitative Results}
\noindent \textbf{Quantitative results.} In our experiments, we adopt the evaluation metrics outlined in Section~\ref{sec4-1}. Multi-line text generation presents unique challenges, including stronger contextual interference, potential mask region overlap, and difficulties in precise positional alignment. Therefore, we provide metrics separately for multi-line (Table~\ref{tab:quan_results1}) and single-line (Table~\ref{tab:quan_results2}) scenarios. The single-line results were obtained by randomly sampling three text instances from the multi-line dataset examples and generating them individually. Although recent approaches such as TextDiffuser, UdiffText, and DreamText have shown promising results, they are limited in two key aspects: they only support single-line text generation, and they are restricted to English. In contrast, the AnyText series support multilingual and multi-line text synthesis, making them more aligned with our setting. Therefore, we select the AnyText series as our primary comparison methods. In addition, we include a detailed comparison with the baseline method Flux.

As shown by the multi-line metrics in Table~\ref{tab:quan_results1}, our method consistently outperforms all existing approaches across all metrics and four benchmark datasets. Even the lightweight LoRA-tuned version surpasses all baselines, demonstrating the effectiveness and adaptability of our approach. When fully trained, our model particularly excels in Chinese text synthesis. On the SeqAcc-Recon metric, it achieves scores of 61.4 for AnyWord(CH) and 64.1 for ReCTS. Furthermore, on the more difficult SeqAcc-Editing metric, its performance on Chinese text, scoring 40.7 on AnyWord(CH) and 37.2 on ReCTS , also substantially exceeds that of baseline methods. Turning to the simpler single-line metrics presented in Table~\ref{tab:quan_results2}, we observe that the multi-line rendering quality does not significantly degrade compared to the single-line results. This further demonstrates our method's accurate positional alignment capability when generating multi-line text. Notably, while the base Flux model is incapable of generating Chinese text, exhibiting zero accuracy on this task, the application of our method unlocks its multilingual text generation capabilities, achieving performance significantly superior to existing approaches.

It is worth noting that the AnyText series of methods do not report performance metrics for the tasks of scene text synthesis (including reconstruction and editing) in their publications. Their publicly available evaluations are limited to the text-to-image generation task, which restricts a comprehensive assessment and comparison of their capabilities. 
To enable a fair and direct comparison, we conducted our own evaluations of the AnyText methods on the specified tasks, ensuring consistent experimental settings.
Furthermore, it is relevant context that scene text synthesis represents an inherently more challenging task compared to standard text-to-image generation. This increased difficulty generally leads to lower quantitative accuracy metrics.



\noindent \textbf{Qualitative results.} Fig.~\ref{visualization} shows multilingual text synthesis results generated by TextFlux under various challenging conditions, such as complex backgrounds, curved text, and handwritten styles. The visualizations demonstrate that TextFlux significantly outperforms existing methods in terms of character accuracy and image fidelity. In most cases, the generated results are nearly indistinguishable from real images. Additionally, we demonstrate the Zero-shot capability in the appendix, which can render languages not included in the training set, such as minority languages.

\begin{figure*}[thbp] 
\centering                
\includegraphics[width=1.0\textwidth]{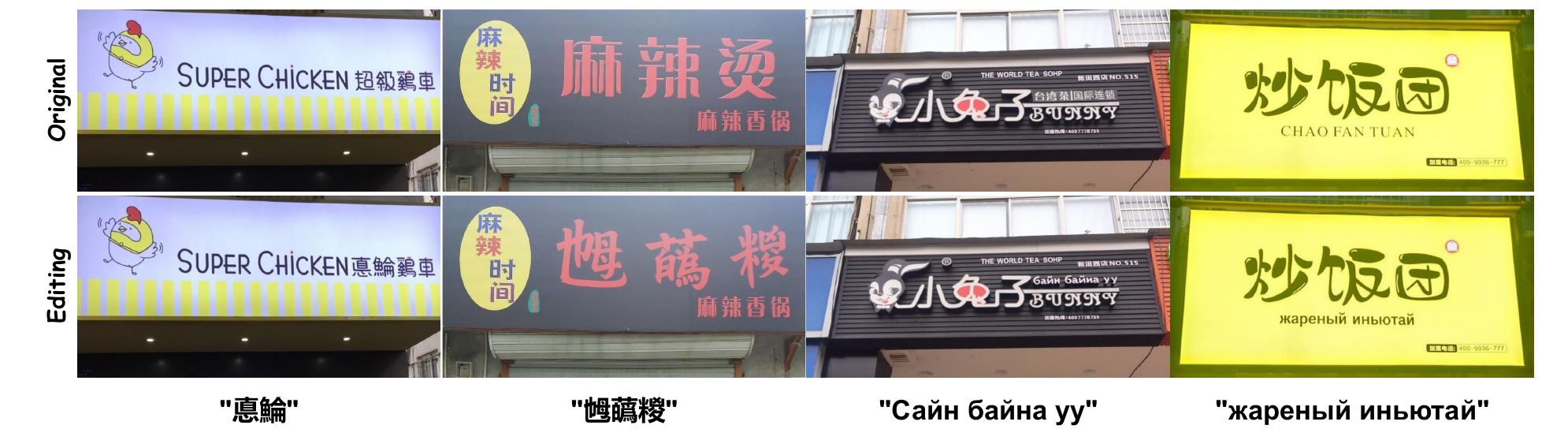}
\caption{\textbf{Zero-shot synthesis of unseen scripts and characters.} The results include rare Chinese characters not present in training and also demonstrate successful generation in Mongolian and Russian, which are languages the model has never seen. These results highlight the generalization ability of TextFlux to novel glyphs.}\label{fig:zero-sot}  
\end{figure*}

We showcase zero-shot visualization results in Fig.~\ref{fig:zero-sot}, , where the model, tasked with generating text unseen during training, consistently demonstrates strong text rendering capabilities. These results suggest that our model does not merely memorize and reproduce trained glyphs but has instead learned a more generalizable and profound capability: to stylistically fuse any given visual glyph reference with the scene context. This generalizable capability is also key to TextFlux's efficiency in handling multilingual text and its strong adaptability to low-resource languages.

\begin{table}[ht]
\centering
\begin{minipage}[t]{0.48\textwidth}
  \centering
  \caption{SeqAcc-Recon results on the ReCTS and TotalText datasets using different training strategies.}
  \begin{tabular}{lcc}
    \toprule
    \textbf{Strategy} & \textbf{ReCTS} & \textbf{TotalText} \\
    \midrule
    No Concat + LoRA     & 5.2   & 29.8 \\
    Concat + No train & 9.2  & 26.2 \\
    Concat + LoRA      & 54.6  & 62.3 \\
    Concat + Full-Param      & \textbf{64.1}  & \textbf{62.9} \\
    \bottomrule
  \end{tabular}
  \label{tab:ablation_strategy}
\end{minipage}
\hfill
\begin{minipage}[t]{0.48\textwidth}
  \centering
  \caption{Evaluating different text encoders based on SeqAcc-Recon results when provided with empty input prompts.}
  \begin{tabular}{cc|cc}
    \toprule
    \textbf{CLIP} & \textbf{T5} & \textbf{ReCTS} & \textbf{TotalText} \\
    \midrule
    \ding{51} & \ding{51} & \textbf{64.1} & \textbf{62.9} \\
    \ding{55} & \ding{51} & 64.0          & 62.7 \\
    \ding{51} & \ding{55} & 63.8          & 55.5 \\
    \ding{55} & \ding{55} & 63.6          & 55.1 \\
    \bottomrule
  \end{tabular}
  \label{tab:ablation_encoder}
\end{minipage}
\end{table}

\subsection{Ablation Study}
\label{ablation-sec}

\begin{figure*}[thbp]
  \centering
  \includegraphics[width=1\textwidth]{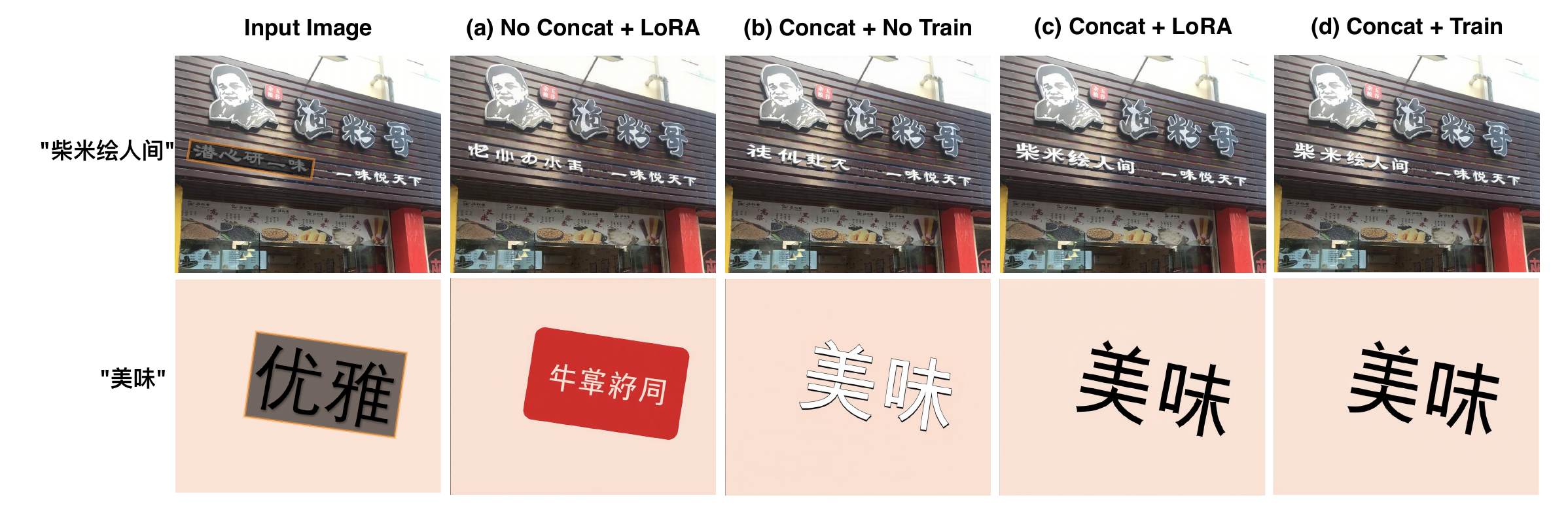}
  \caption{We compare four settings to assess the impact of our glyph concatenation strategy and training schemes on multilingual scene text synthesis.}
  \label{compare_vis_ab}
\end{figure*}

\noindent\textbf{Effectiveness of Concatenation Strategies}
We first analyze the impact of the proposed concatenation strategy and different fine-tuning approaches, with results presented in Table~\ref{tab:ablation_strategy}.
(1) Training directly on the original images without concatenation (No concat + LoRA) achieves a very low sequence accuracy of 5.2\% on ReCTS, failing to generate readable Chinese text (Fig.~\ref{compare_vis_ab}(a)), indicating the base model's limitation.
(2) Using the concatenation strategy without further training (Concat + No train) shows a basic ability to render Chinese (Fig.~\ref{compare_vis_ab}(b), bottom), but fails completely to render recognizable text when the background becomes slightly more complex (Fig.~\ref{compare_vis_ab}(b), top).
(3) Applying LoRA fine-tuning after concatenation (Concat + LoRA) achieves remarkable performance (Fig.~\ref{compare_vis_ab}(c)), highlighting the effectiveness of glyphs as contextual cues even with limited parameter updates.
(4) Full-parameter fine-tuning (Concat + Full-Param) yields the best results, confirming the strategy's scalability and its ability to fully unlock multilingual capabilities.


\noindent\textbf{Impact of Text Encoders on Text Rendering Quality}
We investigate the necessity of text encoders for text rendering in TextFlux, given its primary reliance on visual contextual reasoning. While prior works often emphasize the importance of powerful language modeling in text-to-image generation tasks, we aim to revisit this assumption in the specific context of multilingual visual text generation. Therefore, we train the model by setting the prompts of CLIP or T5 to empty during training to examine the role of textual guidance.
Interestingly, as shown in Table~\ref{tab:ablation_encoder}, our results reveal that removing either the CLIP or T5 encoder individually leads to only marginal changes in rendering performance for non-Latin scripts. For Latin-based languages, removing the T5 encoder results in a 7.4-point performance drop, but the overall rendering quality remains at a high level. These findings suggest that for diffusion models equipped with strong contextual reasoning capabilities, high-quality text rendering can be achieved solely guided by visual context. For future work aiming to further enhance non-Latin text generation capabilities, developing a more efficient text encoder for non-Latin scripts remains a potential research avenue.

\section{Conclusion and Limitations}
In this paper, we propose TextFlux, an OCR-free method that leverages the inherent capabilities of diffusion models to address the intrinsic conflict between generating precise glyph structures and achieving contextually consistent styles. The method not only offers architectural simplicity and significant data efficiency, but also demonstrates strong performance across various aspects, including multilingual support, multi-line editing, complex glyph rendering, and zero-shot generalization. This enables straightforward extension to a wider range of low-resource languages, thereby laying the groundwork for enhanced language accessibility in scene text synthesis. 


However, our method still has some limitations. First, although only approximately 1\% of typical training data is required, training a Flux-based model remains computationally expensive (about four days of training on two 80GB A100 GPUs). Second, the performance of our TextFulx is still unsatisfactory in the task of scene text synthesis for cursive languages, where character representations may differ based on their positions or connections (such as Arabic and Hindi). Third, the proposed framework requires the backbone model to possess strong contextual reasoning capabilities and thus is unsuited for many less-competitive pre-trained models. More details are demonstrated in the supplementary materials. We are planning to address them in our future work.

\bibliographystyle{plain}
\bibliography{sample-base}    

\newpage
\appendix

\section{Visualization of More Generation Results by Our TextFlux}

\begin{figure*}[thbp]
  \centering
  \includegraphics[width=1\textwidth]{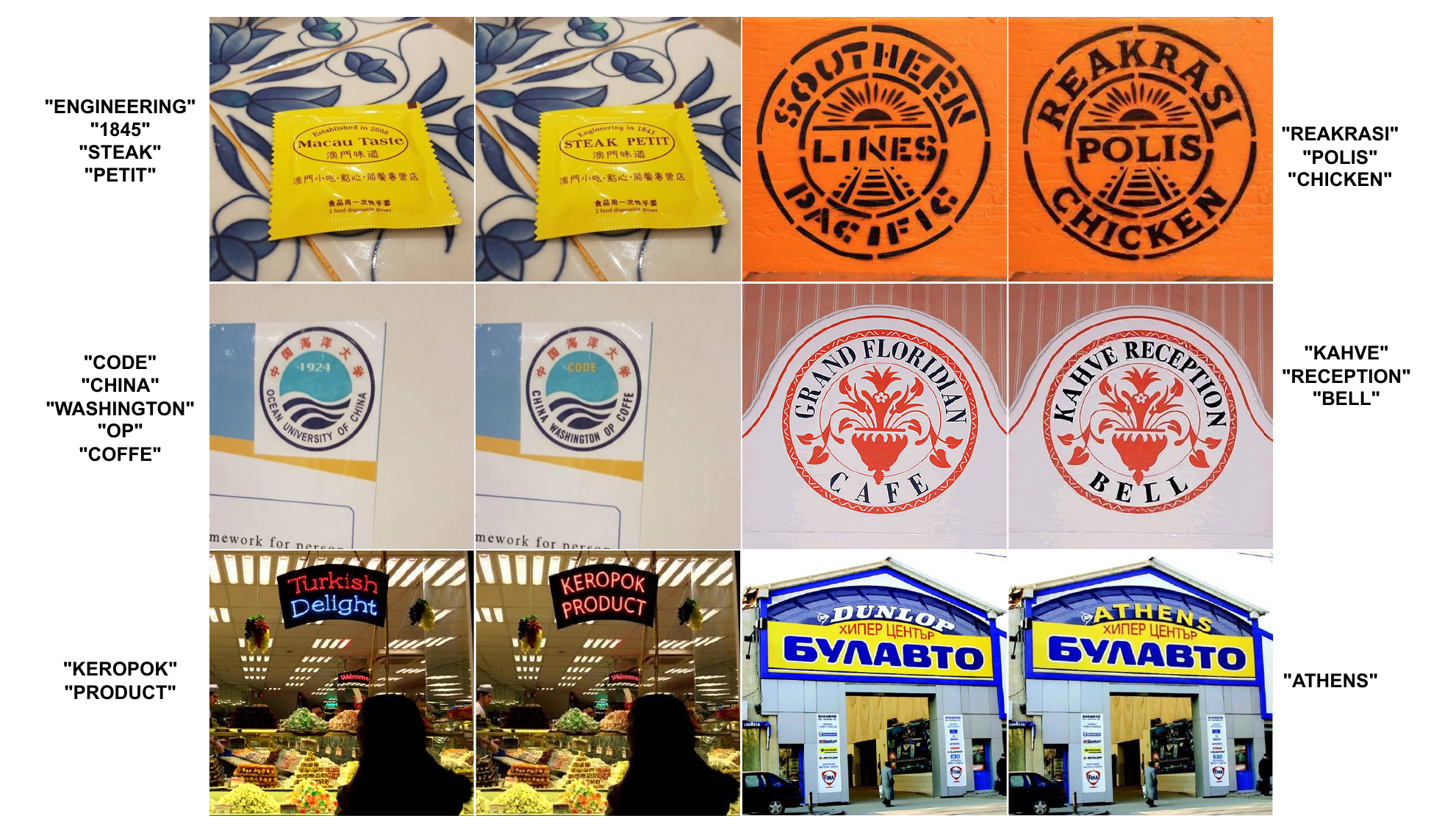}
  \caption{More visualization results of scene text synthesis by TextFlux in English, with a focus on editing multiple lines of text simultaneously.}
  \label{visualization_appendix1}
\end{figure*}
\begin{figure*}[thbp]
  \centering
  \includegraphics[width=1\textwidth]{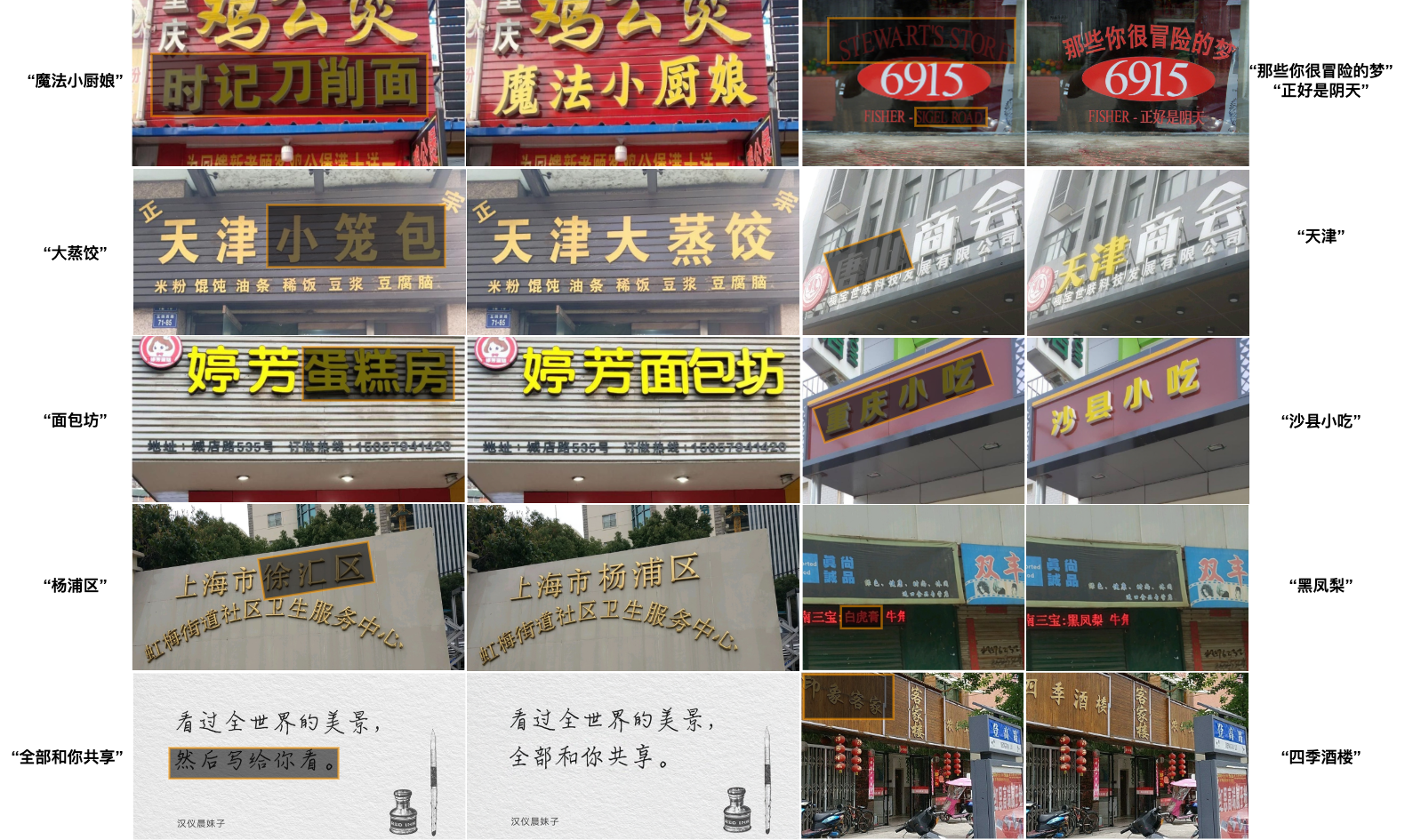}
  \caption{More visualization results of scene text synthesis by TextFlux in Chinese.}
  \label{visualization_appendix2}
\end{figure*}
\begin{figure*}[thbp]
  \centering
  \includegraphics[width=1\textwidth]{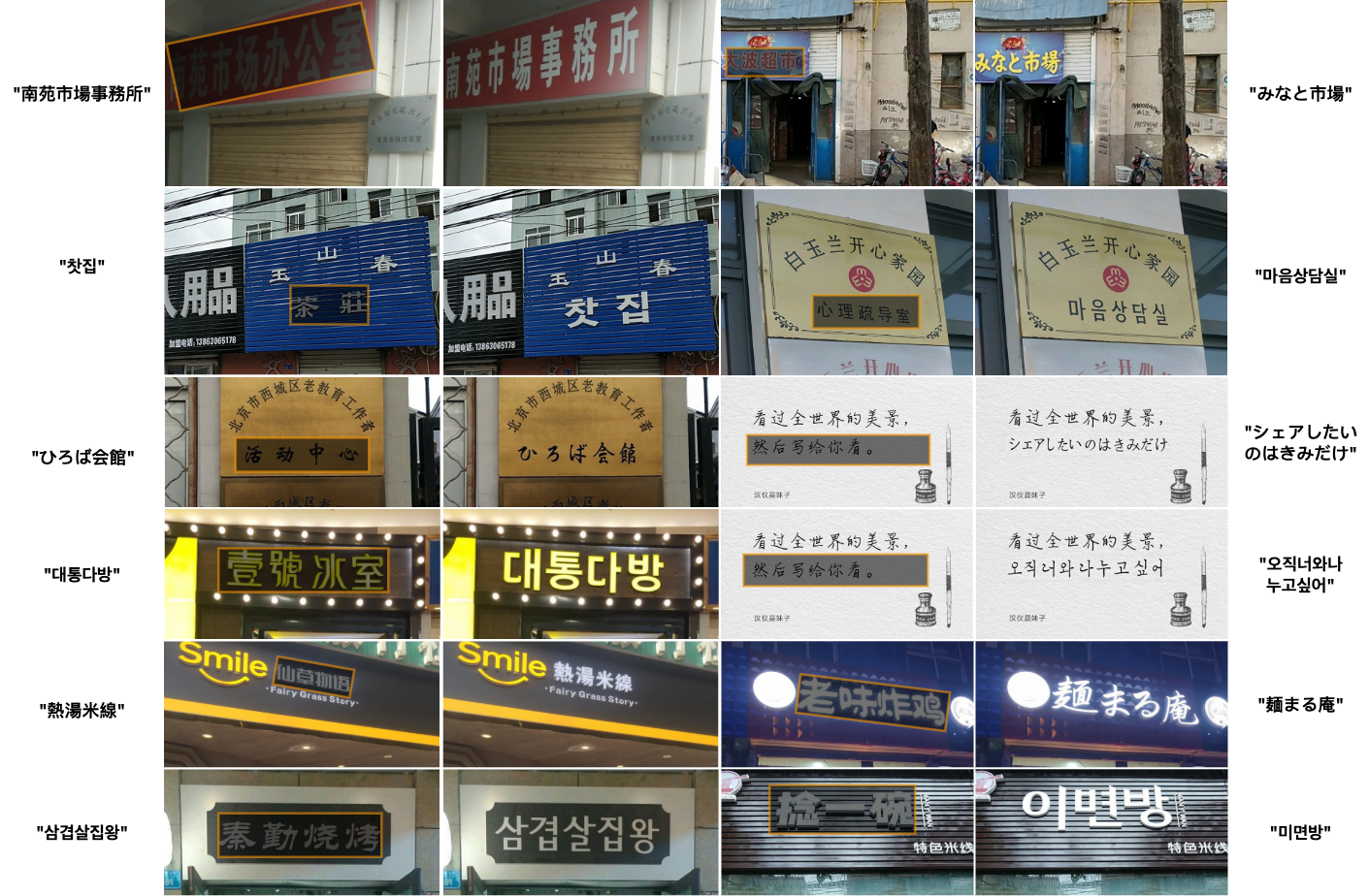}
  \caption{Visualization of multilingual scene text synthesis results in Japanese and Korean.}
  \label{visualization_appendix3}
\end{figure*}
\begin{figure*}[thbp]
  \centering
  \includegraphics[width=1\textwidth]{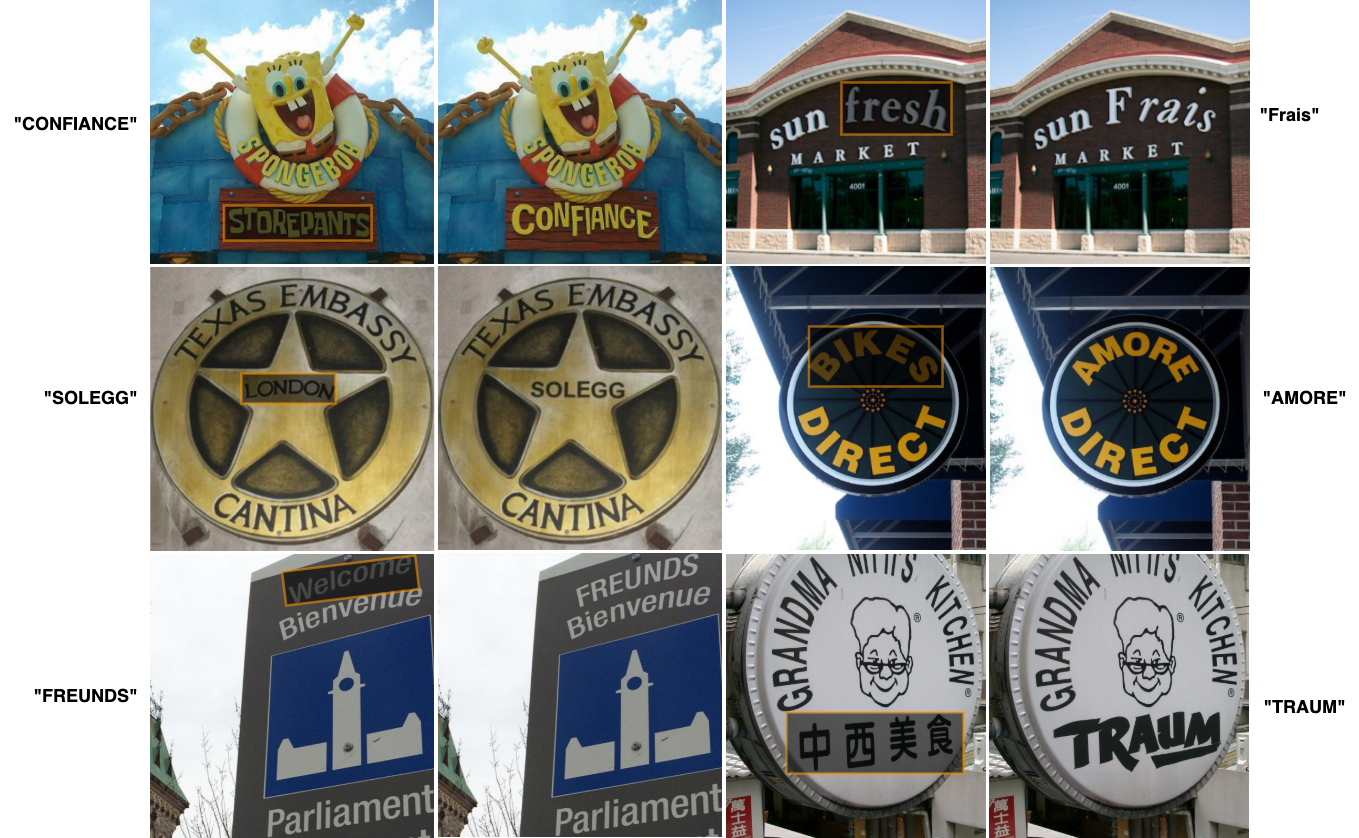}
  \caption{Visualization of multilingual scene text synthesis results in French, Italian, and German. }
  \label{visualization_appendix4}
\end{figure*}

We present additional generation results by TextFlux across a variety of multilingual and complex scene scenarios in Fig.~\ref{visualization_appendix1},~\ref{visualization_appendix2},~\ref{visualization_appendix3},~\ref{visualization_appendix4}. These include challenging cases such as multi-line text editing in English (Fig.~\ref{visualization_appendix1}), scene text synthesis in Chinese (Fig.~\ref{visualization_appendix2}), and few-shot generalization to low-resource scripts such as Japanese, Korean, French, Italian, and German (Fig.~\ref{visualization_appendix3} and Fig.~\ref{visualization_appendix4}).

\section{More Details about Experiments}
This section provides a more detailed breakdown of the quantitative evaluation results on the four benchmark datasets employed in our study: AnyWord (EN), AnyWord (CH), TotalText, and ReCTS. Performance metrics, specifically Sequence Accuracy (SeqAcc), NED, FID, and LPIPS, are reported for both text reconstruction and text editing tasks in Tables~\ref{tab:results_anyword_en}, \ref{tab:results_anyword_ch}, \ref{tab:results_totaltext}, and \ref{tab:results_rects}. All training and evaluation data used in our experiments will be publicly released.


\begin{table}[!t]
    \caption{Additional comparison results on the AnyWord(EN) dataset.}
    \centering
    \resizebox{\linewidth}{!}{
    \begin{tabular}{@{}l|cccc|cccc@{}}
     \toprule[1.5pt]
     \multirow{2}{*}{\textbf{Method}} & \multicolumn{4}{c|}{\textbf{Reconstruction (\%)}} & \multicolumn{4}{c}{\textbf{Editing (\%)}}\\ 
     \cmidrule(lr){2-5} \cmidrule(lr){6-9}
     & \textbf{SeqAcc}$\uparrow$ & \textbf{NED}$\uparrow$ & \textbf{FID}$\downarrow$ & \textbf{LPIPS}$\downarrow$ & \textbf{SeqAcc}$\uparrow$  & \textbf{NED}$\uparrow$  & \textbf{FID}$\downarrow$ & \textbf{LPIPS}$\downarrow$\\
     \midrule[1pt]
     Flux~\cite{blackforestlabs_flux} & 43.0& 59.4& 40.20& 0.1693& 11.6& 23.4& 49.11& 0.2019\\
     AnyText~\cite{tuo2023anytext} & 14.8& 23.8& 32.51& 0.4747& 13.7& 22.7& 27.03& 0.4535\\
     AnyText2~\cite{tuo2024anytext2} & 23.7& 33.2& 33.03& 0.3555& 17.0& 25.9& 28.01& 0.3205\\
     \midrule[1pt]
     TextFlux(LoRA) & \underline{76.7}& \underline{89.1}& \underline{18.85}& \underline{0.0933}& \underline{61.1}& \underline{75.8}& \textbf{24.81}& \textbf{0.1374}\\
     TextFlux & \textbf{77.3}& \textbf{90.2}& \textbf{18.76}& \textbf{0.0933}& \textbf{63.8}& \textbf{78.9}& \underline{25.62}& \underline{0.1379}\\
     \bottomrule[1.5pt]
    \end{tabular}}
    \label{tab:results_anyword_en}
\end{table}

\begin{table}[!t]
    \caption{Additional comparison results on the AnyWord(CH) dataset.}
    \centering
    \resizebox{\linewidth}{!}{
    \begin{tabular}{@{}l|cccc|cccc@{}}
     \toprule[1.5pt]
     \multirow{2}{*}{\textbf{Method}} & \multicolumn{4}{c|}{\textbf{Reconstruction (\%)}} & \multicolumn{4}{c}{\textbf{Editing (\%)}}\\ 
     \cmidrule(lr){2-5} \cmidrule(lr){6-9}
     & \textbf{SeqAcc}$\uparrow$ & \textbf{NED}$\uparrow$ & \textbf{FID}$\downarrow$ & \textbf{LPIPS}$\downarrow$ & \textbf{SeqAcc}$\uparrow$  & \textbf{NED}$\uparrow$  & \textbf{FID}$\downarrow$ & \textbf{LPIPS}$\downarrow$ \\
     \midrule[1pt]
     Flux~\cite{blackforestlabs_flux} & 9.3& 13.6& 24.22& 0.1406& 0.0& 0.0& 29.24& 0.1496\\
     AnyText~\cite{tuo2023anytext} & 24.1& 34.1& 33.59& 0.7766& 19.2& 30.8& 32.36& 0.7784\\
     AnyText2~\cite{tuo2024anytext2} & 28.1& 38.1& 27.88& 0.5945& 24.2& 35.5& 27.12& 0.5969\\
     \midrule[1pt]
     TextFlux(LoRA) & \underline{50.8}& \underline{77.5}& \underline{15.69}& \underline{0.0732}& \underline{32.8}& \underline{57.7}& \underline{21.09} & \underline{0.0995}\\
     TextFlux & \textbf{61.4}& \textbf{82.0}& \textbf{14.41}& \textbf{0.0695}& \textbf{40.7}& \textbf{66.4}& \textbf{19.79}& \textbf{0.0993}\\
     \bottomrule[1.5pt]
    \end{tabular}}
    \label{tab:results_anyword_ch}
\end{table}

\begin{table}[!t]
    \caption{Additional comparison results on the TotalText dataset.}
    \centering
    \resizebox{\linewidth}{!}{
    \begin{tabular}{@{}l|cccc|cccc@{}}
     \toprule[1.5pt]
     \multirow{2}{*}{\textbf{Method}} & \multicolumn{4}{c|}{\textbf{Reconstruction (\%)}} & \multicolumn{4}{c}{\textbf{Editing (\%)}}\\ 
     \cmidrule(lr){2-5} \cmidrule(lr){6-9}
     & \textbf{SeqAcc}$\uparrow $& \textbf{NED}$\uparrow$ & \textbf{FID}$\downarrow$ & \textbf{LPIPS}$\downarrow$ & \textbf{SeqAcc}$\uparrow$  & \textbf{NED}$\uparrow$  & \textbf{FID}$\downarrow$ & \textbf{LPIPS}$\downarrow$ \\
     \midrule[1pt]
     Flux~\cite{blackforestlabs_flux} & 29.5& 45.5& 17.89& 0.0701& 11.5& 26.9& 21.12& 0.0798\\
     AnyText~\cite{tuo2023anytext} & 6.5& 16.7& 41.39& 0.3718& 4.6& 13.6& 40.55& 0.3537\\
     AnyText2~\cite{tuo2024anytext2} & 15.5& 27.9& 33.48& 0.2715& 15.0& 25.3& 32.47& 0.2413\\
     \midrule[1pt]
     TextFlux(LoRA) & \underline{62.3}& \underline{77.2}& \underline{12.11}& \underline{0.0556}& \underline{35.4}& \underline{55.4}& \underline{16.58}& \textbf{0.0710}\\
     TextFlux & \textbf{62.9}& \textbf{78.7}& \textbf{11.72}& \textbf{0.0554}& \textbf{36.2}& \textbf{57.6}& \textbf{16.26}& \underline{0.0714}\\
     \bottomrule[1.5pt]
    \end{tabular}}
    \label{tab:results_totaltext}
\end{table}

\begin{table}[!t]
    \caption{Additional comparison results on the ReCTS dataset.}
    \centering
    \resizebox{\linewidth}{!}{
    \begin{tabular}{@{}l|cccc|cccc@{}}
     \toprule[1.5pt]
     \multirow{2}{*}{\textbf{Method}} & \multicolumn{4}{c|}{\textbf{Reconstruction (\%)}} & \multicolumn{4}{c}{\textbf{Editing (\%)}}\\ 
     \cmidrule(lr){2-5} \cmidrule(lr){6-9}
     & \textbf{SeqAcc}$\uparrow$ & \textbf{NED}$\uparrow$ & \textbf{FID}$\downarrow$ & \textbf{LPIPS}$\downarrow$ & \textbf{SeqAcc}$\uparrow$  & \textbf{NED}$\uparrow$  & \textbf{FID}$\downarrow$ & \textbf{LPIPS}$\downarrow$ \\
     \midrule[1pt]
     Flux~\cite{blackforestlabs_flux} & 4.8& 8.7& 18.29& 0.1432& 0.0& 0.0& 19.38& 0.1439\\
     AnyText~\cite{tuo2023anytext} & 20.6& 29.4& 22.18& 0.4091& 18.5& 25.7& 22.96& 0.4099\\
     AnyText2~\cite{tuo2024anytext2} & 25.2& 34.2& 21.66& 0.3049& 23.6& 29.9& 21.84& 0.3059\\
     \midrule[1pt]
     TextFlux(LoRA) & \underline{56.6}& \underline{74.8}& \underline{12.09}& \underline{0.1038}& \underline{32.1}& \underline{53.4}& \underline{14.15}& \underline{0.1274}\\
     TextFlux & \textbf{64.1}& \textbf{79.6}& \textbf{11.02}& \textbf{0.0975}& \textbf{37.2}& \textbf{58.9}& \textbf{13.41}& \textbf{0.1258}\\
     \bottomrule[1.5pt]
    \end{tabular}}
    \label{tab:results_rects}
\end{table}

\section{Qualitative Comparison with Flux on General Inpainting Tasks}
\begin{figure*}[thbp]
  \centering
  \includegraphics[width=1\textwidth]{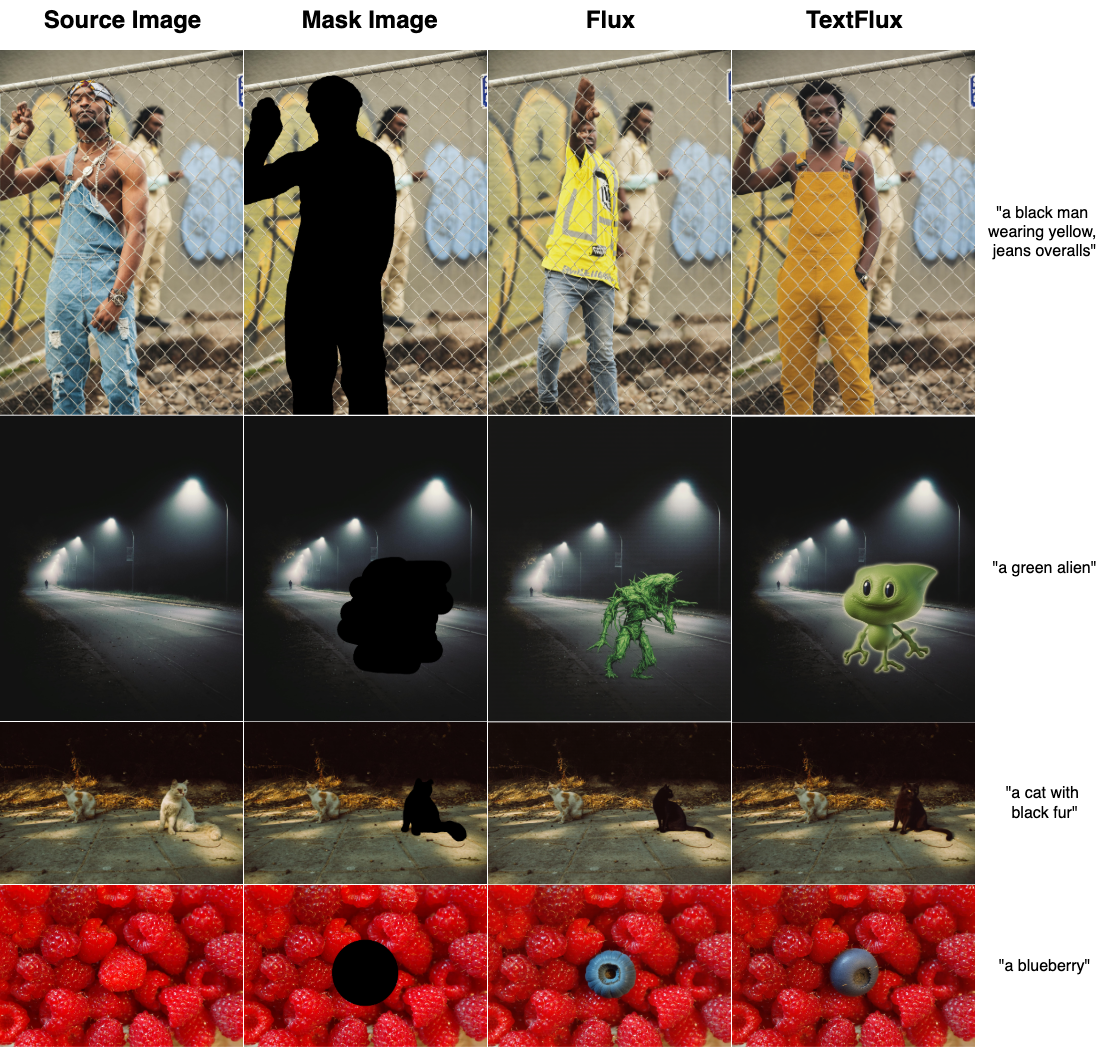}
  \caption{Visualization of general inpainting tasks using Flux and TextFlux under the same prompt and mask conditions. Prompt texts are shown on the right. }
  \label{visualization_appendix_general}
\end{figure*}

We selected the first few sample images from the evaluation benchmark in the original Flux~\cite{blackforestlabs_flux} codebase and compared the performance of TextFlux and the original Flux under the same mask and prompt conditions. As shown in Fig.~\ref{visualization_appendix_general}, the results show that TextFlux achieves almost the same generation capability as Flux in handling various types of inpainting tasks.

Specifically, in the human editing task, TextFlux can accurately understand the prompt “a black man wearing yellow, jeans overalls” and perform a natural and reasonable clothing replacement. The generated result even surpasses the original Flux in terms of visual style and background consistency. In the reconstruction of imaginary objects (such as “a green alien”), detail restoration (such as replacing with “a blueberry”), and animal editing tasks (such as “a cat with black fur”), the generation quality of TextFlux is also comparable to Flux.

These results show that although TextFlux is designed for text image synthesis tasks, its adaptation ability in general inpainting scenarios is still preserved. This lays a foundation for extending the method in this paper to broader multi-modal image editing tasks in the future.

\section{Controllability via Prompt Modification}
\begin{figure*}[thbp]
  \centering
  \includegraphics[width=1\textwidth]{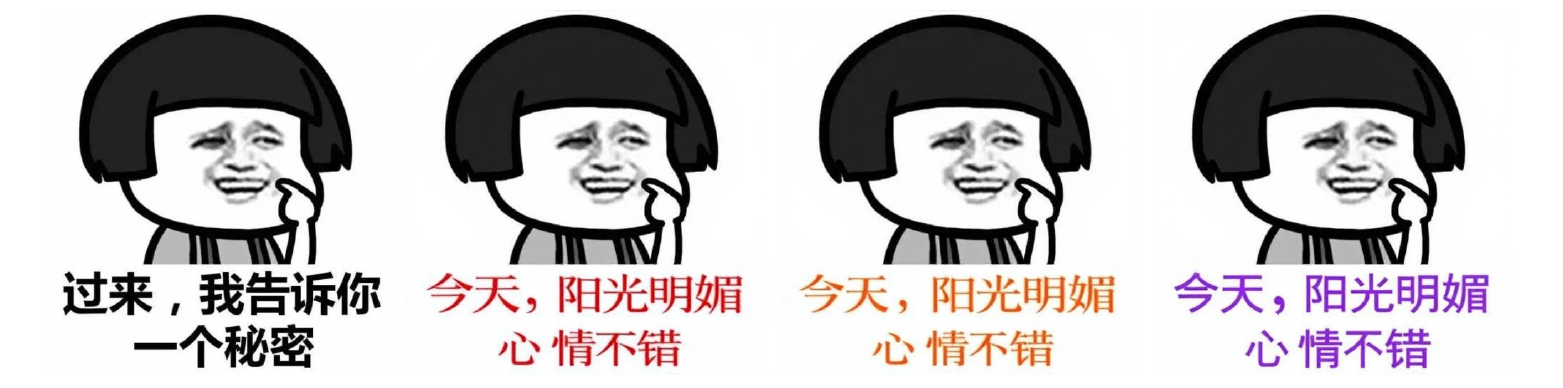}
  \caption{TextFlux responds to different color prompts such as ``The generated text should be red/orange/purple''.}
  \label{visualization_appendix_color}
\end{figure*}

Although TextFlux uses a standardized descriptive prompt template during training to clarify the roles of each part in the concatenated input image, we further investigate whether it is possible to achieve a certain degree of controllability at the inference stage by simply modifying the prompt. Specifically, we add a sentence at the end of the original prompt: “The generated text should be {color}.”, where {color} can be selected as needed. According to the visualization results in Fig.~\ref{visualization_appendix_color}, TextFlux still retains some response ability to such simple attributes, showing basic controllability.

\section{Further Limitations and Discussion}
\begin{figure*}[thbp]
  \centering
  \includegraphics[width=1\textwidth]{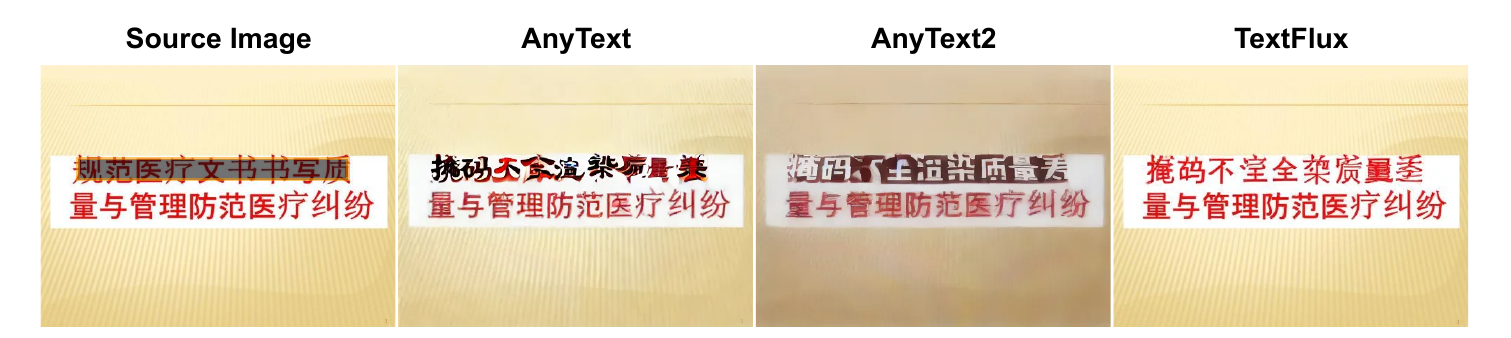}
  \caption{Impact of slight character cropping on synthesis quality. When the mask slightly cuts into the line of text, all methods show a significant drop in rendering quality.}
  \label{visualization_appendix_mask}
\end{figure*}

\noindent\textbf{Impact of Mask Coverage on Synthesis Quality.}
We further analyze the impact of mask coverage on synthesis quality, which is crucial in real-world applications. Our quantitative evaluations are usually based on masks derived from tight bounding box annotations in the dataset. However, we observe that if these masks do not fully cover the target text region (for example, slightly crop characters), they may lead to severe visual artifacts and rendering errors (see Fig.~\ref{visualization_appendix_mask}). This issue is not unique to TextFlux and can also be observed in methods like AnyText2~\cite{tuo2024anytext2}. In contrast, real users often create looser masks during editing tasks, leaving some padding around the text. When using such masks, TextFlux and other methods tend to produce more coherent and complete visual results. This suggests that although evaluation with tight masks is the standard practice, it may not fully reflect the more robust performance that can be achieved in practical usage when more tolerant masks are applied.

\begin{figure*}[thbp]
  \centering
  \includegraphics[width=1\textwidth]{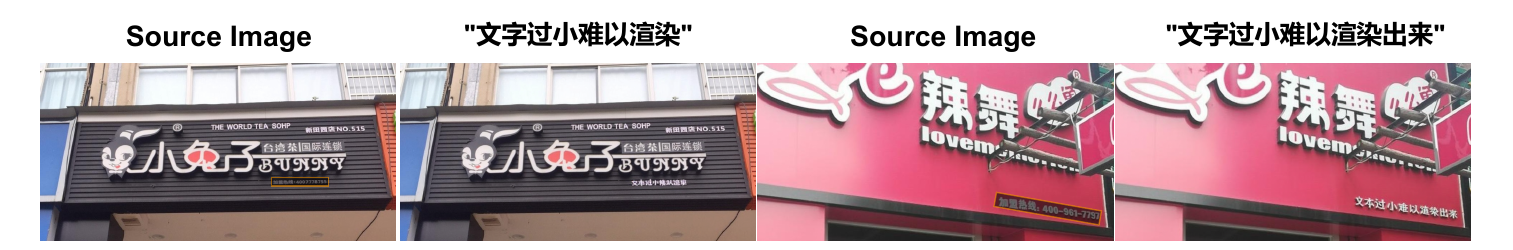}
  \caption{Difficulty in rendering extremely small text. When the target text is too small, the model struggles to preserve fine details, often leading to blurry or illegible results.}
  \label{visualization_appendix_small_text}
\end{figure*}

\noindent\textbf{Challenges in Rendering Extremely Small Text.}
Another challenge is the synthesis of extremely small text. TextFlux relies on provided visual glyph templates to guide the fine-grained appearance of characters. When the target text is very small, the resolution of the glyph image becomes low, making it difficult for the model to preserve fine character details during the VAE encoding-decoding and the subsequent diffusion-based stylization process. As shown in Fig.~\ref{visualization_appendix_small_text}, although TextFlux still tries to render the text, the readability and structural integrity of very small characters can be affected, resulting in blurred or distorted glyphs. This indicates that the quality of visual glyph guidance largely depends on whether the input glyph contains enough pixel information to clearly represent its structure.

\begin{figure*}[thbp]
  \centering
  \includegraphics[width=1\textwidth]{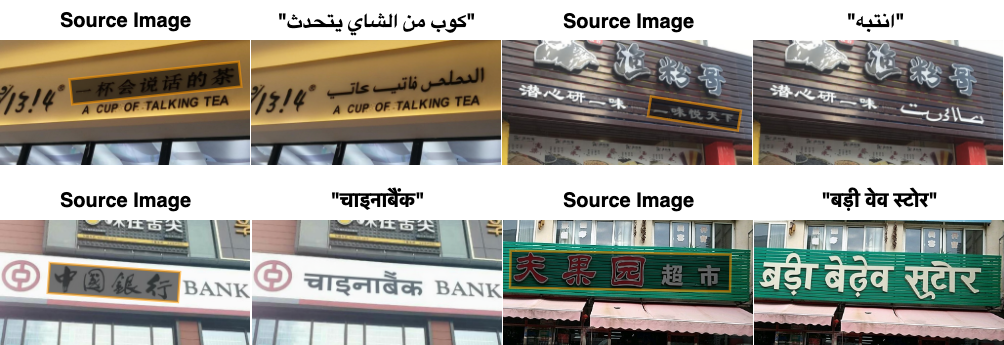}
  \caption{Limitation in rendering cursive scripts. For highly connected writing systems like Arabic (first row) and Hindi (second row), TextFlux struggles to reproduce accurate character connections and shapes, leading to structural distortions in the generated text.}
  \label{visualization_appendix_cursive}
\end{figure*}

\noindent\textbf{Difficulties with Cursive Scripts.}
Generating text in highly cursive writing systems (such as Arabic or Hindi) presents unique challenges. Although corresponding glyph templates are provided, the appearance of isolated characters is very different from how they appear when rendered in connected forms. The model has difficulty accurately learning this mapping. As shown in Fig.~\ref{visualization_appendix_cursive}, TextFlux can roughly reproduce the writing direction and general shapes of these scripts, but there are still significant limitations in finer details, making it hard to fully meet practical requirements. This challenge comes from the fact that training inputs are isolated glyphs, while the desired outputs are connected cursive text, which involves a complex visual mapping. In the future, for such scripts, it may be necessary to design specific glyph rendering strategies that support cursive structures.

\subsection{Broader Impact and Ethical Considerations}
The high realism and fidelity achieved by TextFlux in synthesizing text within scenes is a core research goal of our work, but it also reveals potential social risks and ethical concerns. If misused, the ability to seamlessly and convincingly modify text in images could be used to generate misleading or malicious content. Possible misuse scenarios include: (1) altering existing text in images to fabricate information or evidence, such as modifying signs, screenshots, or documents to support false narratives; (2) realistically modifying identity cards, certificates, or other official documents to forge identity or alter sensitive information; (3) creating more deceptive forgeries or phishing materials.

Although the main purpose of this work is to support creative applications, improve accessibility, and advance the fundamental research on controllable image synthesis, we are aware that this technology may have dual-use characteristics. As with other powerful generative AI systems, its potential benefits must be weighed against possible risks of misuse. We encourage the research community to pay close attention to these issues and contribute to the development of protective measures, such as methods for detecting text modification in images, establishing ethical guidelines for the use of such technologies, and exploring techniques like digital watermarking to identify synthetic content. Our goal is to promote the progress of scene text synthesis in a responsible manner.

\end{document}